\documentclass{article}

\usepackage{PRIMEarxiv}

\usepackage[utf8]{inputenc} 
\usepackage[T1]{fontenc}    
\usepackage{hyperref}       
\usepackage{url}            
\usepackage{booktabs}       
\usepackage{amsfonts}       
\usepackage{nicefrac}       
\usepackage{microtype}      
\usepackage{lipsum}
\usepackage{fancyhdr}       
\usepackage{graphicx}       
\graphicspath{{media/}}     
\usepackage{subcaption} 
\usepackage{algorithm}
\usepackage{algorithmic}
\usepackage{changepage}
\usepackage{amsmath}
 \usepackage{multirow}
\title{Lightweight Vision Transformer with Window and Spatial Attention for Food Image Classification}

\author{
  Xinle Gao, Linghui Ye \\
  School of Artificial Intelligence and Computer Science \\
  Jiangnan University \\
  Wuxi, 214122, China.\\
   \And
  Zhiyong Xiao* \\
  School of Artificial Intelligence and Computer Science \\
  Jiangnan University \\
  Wuxi, 214122, China.\\
  \texttt{zhiyong.xiao@jiangnan.edu.cn} \\
}

\begin{document}
\maketitle

\begin{abstract}
\label{}
With the rapid development of society and continuous advances in science and technology, the food industry increasingly demands higher production quality and efficiency. Food image classification plays a vital role in enabling automated quality control on production lines, supporting food safety supervision, and promoting intelligent agricultural production. However, this task faces challenges due to the large number of parameters and high computational complexity of Vision Transformer models. To address these issues, we propose a lightweight food image classification algorithm that integrates a Window Multi-Head Attention Mechanism (WMHAM) and a Spatial Attention Mechanism (SAM). The WMHAM reduces computational cost by capturing local and global contextual features through efficient window partitioning, while the SAM adaptively emphasizes key spatial regions to improve discriminative feature representation. Experiments conducted on the Food-101 and Vireo Food-172 datasets demonstrate that our model achieves accuracies of 95.24\% and 94.33\%, respectively, while significantly reducing parameters and FLOPs compared with baseline methods. These results confirm that the proposed approach achieves an effective balance between computational efficiency and classification performance, making it well-suited for deployment in resource-constrained environments.
\end{abstract}


\section{Introduction}
With the rapid development of society and continuous advances in science and technology, the food industry faces increasingly stringent demands for high-quality and efficient production \cite{ciocca2017food}. In this context, food image classification has become a critical technology, enabling automated quality inspection on production lines, strengthening food safety supervision, and accelerating the intelligent transformation of agricultural production.

Despite its importance, food image classification remains a challenging task due to the diversity and complexity of food categories \cite{eccv2018foodexpansion}. Foods within the same category often share highly similar attributes, such as color and texture, while exhibiting variations in shape, perspective, and fine-grained details. These similarities hinder precise subcategory classification, while differences can lead to confusion in feature learning, increasing classification complexity. In addition to advances in food image classification, research on hyperspectral image (HSI) processing has provided valuable insights into efficient feature extraction and lightweight model design \cite{RSL2014,Xiao2014,Liu2017}.

In recent years, deep learning has significantly advanced food image classification. Convolutional Neural Networks (CNNs) have played a central role by enabling robust feature extraction and effectively addressing the challenges of high intra-class similarity and inter-class variability \cite{meyers2015im2calories}. Several CNN-based approaches have been proposed, ranging from region-based feature extraction to lightweight architectures tailored for resource efficiency. Wei and Wang \cite{wei2022foodimage} proposed a model employing the Faster R-CNN architecture for food region detection in dietary images, followed by CNN-based feature extraction from detected food regions. Yonis et al. \cite{gulzar2023fruit} implemented a modified version of MobileNet V2, termed TL-MobileNet V2, for fruit classification. CNN architectures also achieved remarkable advancements in medical image analysis \cite{Xiao2019,Qian2020,CMPB2021,BSPC2021,Liu2022}. However, CNNs are inherently limited in their ability to capture long-range dependencies, which constrains their performance in complex classification tasks.

The introduction of Vision Transformers (ViTs) has provided a new perspective for food image classification \cite{dosovitskiy2021vit}. Using self-attention mechanisms, ViTs model global semantic dependencies across image patches, leading to improved representation learning . The Swin Transformer further improves efficiency by employing shifted window attention, reducing parameter counts and computational complexity while maintaining competitive performance in vision tasks \cite{han2022survey}. 
Recently, several studies have further advanced food image classification by addressing fine-grained challenges and improving model generalization. For instance, Gao et al. successfully tackled the intricate task of classifying food images with highly similar morphologies by employing sophisticated data augmentation and feature enhancement strategies within the Vision Transformer framework (AlsmViT) \cite{Xiao2024FoodViT}. Similarly, Xiao et al. introduced a deep convolutional module that, when integrated with the global representations of a Swin Transformer, significantly enhanced both local and global feature representations \cite{Xiao2024FoodSwinViT}. Xiao et al. proposed a high-accuracy food image recognition model (FoodCSWin) that enhances dietary assessment by combining diffusion-based data augmentation and local feature enhancement to handle visually diverse but nutritionally similar foods \cite{XIAO2025FoodCSWin}. Beyond transformer-based approaches, researchers have also explored hybrid optimization strategies. By translating the hierarchical parallelism and evolutionary optimization philosophy of hybrid parallel genetic algorithms into food classification, notable improvements have been achieved, particularly in handling class-imbalanced and fine-grained recognition scenarios \cite{HPGA2021}. 

However, the structural characteristics of ViTs present two key limitations: (1) the large number of parameters and computational overhead significantly increase training and inference costs, and (2) their resource-intensive nature restricts deployment in real-world applications, particularly on mobile or embedded devices with limited computing power. These challenges underscore the need for lightweight transformer-based models that balance classification performance with computational efficiency. 

In related domains, recent studies have also highlighted the effectiveness of combining CNNs and Transformers for medical image analysis tasks. Xiao et al. proposed a semi-supervised optimization strategy based on the teacher–student paradigm that integrates the strengths of CNNs and Transformers, significantly improving the utilization of large volumes of unlabeled medical images and enhancing segmentation accuracy \cite{XIAO2022CMPB}. Ji et al. introduced an MLP-based model that employs matrix decomposition and rolling tensor operations for skin lesion segmentation, effectively replacing the self-attention mechanism in Transformers. This approach not only achieved superior performance with fewer parameters but also demonstrated that revisiting traditional methods can yield innovative breakthroughs in deep learning \cite{RMMLPBSPC2023}. Furthermore, the adaptability of hybrid architectures has been illustrated through the Light3DHS model, which combines CNN and ViT components for 3D hippocampal structure segmentation. Light3DHS improved segmentation accuracy while reducing computational cost, offering valuable contributions to brain disease research and clinical applications \cite{XIAO2024Light3DHS}.
Hybrid CNN-Transformer architectures have also been widely applied in other fields \cite{BEL2024,FoodSeg2025,XIAO2025pestclassification,Wang2025}. 
More recently, lightweight architectures have emerged as a promising direction. For example, a fine-grained food recognition model integrating an FIP-Attention module for modeling complex ingredient–dish relationships, together with an FCR-Classifier for refining texture–color features, has achieved state-of-the-art performance across multiple benchmarks and demonstrated practical applicability in mobile dietary monitoring applications \cite{XIAO2025FGFoodNet}. 

Although hybrid CNN-Transformer architectures have shown promising results in food classification, medical imaging, and hyperspectral analysis, they are not without limitations. The integration of convolutional modules and self-attention layers often increases architectural complexity, leading to larger parameter counts and higher FLOPs \cite{liu2021swin}. 

To address these issues, we propose a lightweight food image classification algorithm that integrates a Window Multi-Head Attention Mechanism (WMHAM) with a Spatial Attention Mechanism (SAM). The WMHAM partitions input features into nonoverlapping local windows, reducing computational cost while retaining global contextual information through interwindow interactions. The SAM adaptively emphasizes discriminative spatial regions, enhancing the focus of the model on key features without increasing the number of parameters. Together, these mechanisms effectively reduce the complexity of the model while improving the classification performance.

We evaluate our approach on two widely used benchmark datasets, Food-101 and Vireo Food-172, achieving classification accuracies of 95.24\% and 94.33\%, respectively. Compared with existing transformer-based methods, our model significantly reduces parameters and FLOPs, while maintaining or improving accuracy, precision, recall, and F1-score. These results demonstrate that the proposed approach achieves an effective balance between efficiency and performance, making it well-suited for resource-constrained applications in real-world food industry scenarios.

The contributions of this paper are threefold:

We propose a lightweight Vision Transformer-based food image classification algorithm that integrates WMHAM and SAM to reduce computational cost and improve performance.

We conduct extensive ablation and comparative experiments to validate the effectiveness of the proposed mechanisms.

We demonstrate that our approach achieves state-of-the-art performance with significantly fewer parameters and FLOPs, offering a practical solution for deployment in intelligent food processing and safety monitoring systems.

\section{Methods}

We propose a lightweight food image classification algorithm that integrates Window Multi-Head Attention (WMHA) and a Spatial Attention Module (SAM) to balance computational efficiency with classification accuracy. The overall architecture is illustrated in Figure \ref{fig2-1}.

The core idea is to replace the standard multi-head attention in Vision Transformer with a window-based mechanism, while embedding spatial attention in residual form within the feed-forward network. This design reduces parameter count, mitigates computational complexity, and enhances feature representation.

The proposed method primarily consists of window attention and spatial attention mechanisms. The window attention mechanism combines the advantages of relative positional encoding and multi-head attention, enabling more precise capture of relative positional relationships within sequences while enhancing model expressiveness. The relative positional encoding adopts a more compact encoding scheme that maps multiple relative positions to shared encodings, thereby reducing required parameters. In the multi-head attention mechanism, certain parameters such as weight matrices can be shared across different attention heads. Through parameter sharing, we reduce the number of independent parameters per head, thus decreasing total parameter count.

The spatial attention mechanism is embedded in residual form within Vision Transformer's feed-forward neural network. This approach better extracts and leverages critical features to improve model accuracy while enhancing robustness to input variations and boosting performance without increasing parameter count. Ablation experiments and comparative trials demonstrate that our lightweight food image classification algorithm with relative positional encoding attention successfully reduces parameter count in the second model component while improving evaluation metrics including accuracy, precision, and recall. Our model achieves classification accuracies of 95.24\% on Food-101 and 94.33\% on Vireo Food-172 datasets, demonstrating excellent computational efficiency and classification performances.

\label{}
\subsection{Overall architecture of the approach}
\begin{figure*}[!ht]
	\centering	
	\includegraphics[width=16.29cm,height=9cm]{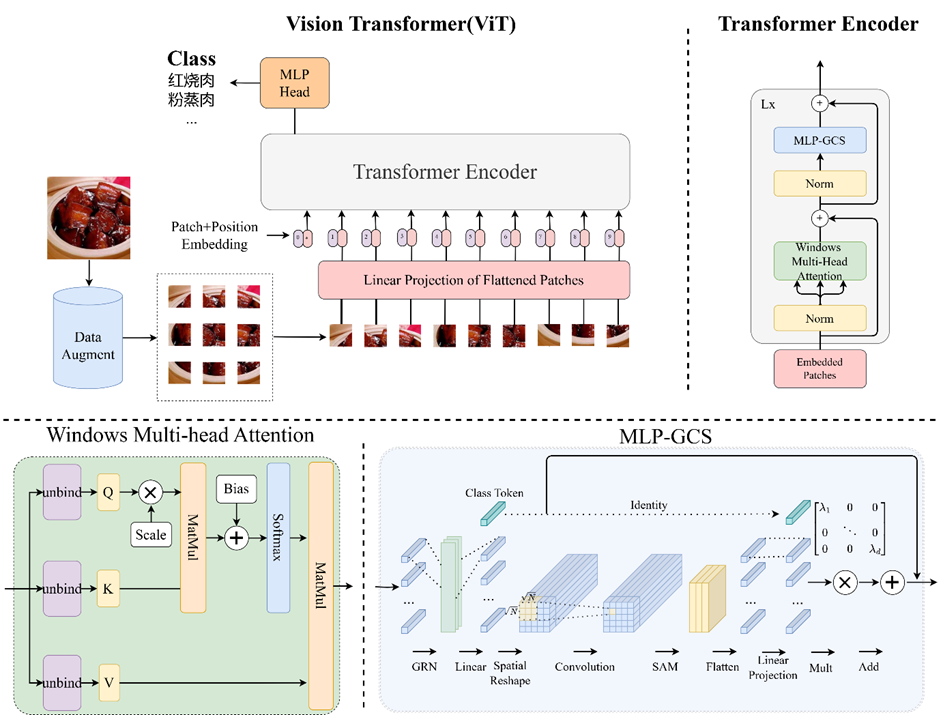}
	\caption{Diagram of a lightweight food image classification algorithm based on window multi-head attention.}
	\label{fig2-1}
\end{figure*}

The structure diagram of the lightweight food image classification algorithm based on window multi-head attention is shown in Figure \ref{fig2-1}. Firstly, we replace the original multi-head attention mechanism with the Window Multi-Head Attention Mechanism. The traditional multi-head attention mechanism has a large computational cost when processing global information, especially on high-resolution images, where the computational complexity increases significantly \cite{vaswani2017attention}. The window multi-head attention mechanism reduces the computational cost by dividing the input feature map into multiple local windows and independently calculating the attention weights within each window. This mechanism not only captures local features, but also retains global context information through interwindow information interaction \cite{liu2021swin}, thus reducing computational costs while maintaining high classification performance. Specifically, the window multi-head attention mechanism divides the input feature map into multiple non-overlapping windows, and the pixels within each window only calculate attention with other pixels within the same window. This localized calculation method greatly reduces computational complexity, especially when processing high-resolution images, significantly reducing memory usage and computation time. In addition, to maintain the transmission of global information, the window multi-head attention mechanism also introduces an inter-window information interaction mechanism, ensuring that the model can capture global context information through cross-window attention calculation. This design enables the model to focus on local details while maintaining an understanding of the global structure when processing complex scenes, thereby improving classification accuracy.

Secondly, we introduced a Spatial Attention Module after the convolution operation in the MLP-GC model. The Spatial Attention Module can effectively enhance the model's focus on key regions and suppress the interference of irrelevant background information by adaptively learning the importance of each spatial position in the feature map. Specifically, the Spatial Attention Module first performs global average pooling and max pooling on the feature map output by the convolution\cite{woo2018cbam}, then concatenates the results and generates spatial attention weights through a convolutional layer. These weights are multiplied by the original feature map to highlight the feature representation of important regions. Finally, we flatten the weighted feature map to facilitate classification by the subsequent fully connected layer. The introduction of the Spatial Attention Module enables the model to handle feature information of different scales more flexibly. Through global average pooling and max pooling operations, the model can capture global and local significant information in the feature map, thereby generating more accurate spatial attention weights. These weights can effectively enhance the model's focus on key regions and suppress the interference of background noise, thereby improving classification accuracy. Additionally, the computational complexity of the Spatial Attention Module is relatively low and does not significantly increase the model's computational burden. Therefore, while maintaining efficient computation, it can effectively enhance the model's performance.

By introducing the window multi-head attention mechanism and the Spatial Attention Module, we not only reduced the model's parameter count and computational complexity but also significantly improved its classification accuracy. These improvements enable the model to handle complex food images more effectively.

\subsection{Spatial attention}
The spatial attention mechanism can be viewed as an adaptive spatial region selection mechanism. In information processing systems, whether in the human brain or computers, the ability to perceive and process information in specific regions is enhanced through focused attention. This mechanism enables systems to more rapidly and accurately capture environmental information, facilitating subsequent learning, reasoning, and actions. In deep learning, the spatial attention mechanism improves model performance by incorporating attention mechanisms into architectures like Convolutional Neural Networks (CNNs), allowing models to adaptively focus on the most critical spatial regions in input data \cite{wang2017residual}. This is typically achieved by computing weight values for different spatial locations in the input data, where higher weights indicate greater importance of those locations for the current task.
 
\begin{figure*}[!ht]
	\centering	
	\includegraphics[width=16.29cm,height=9cm]{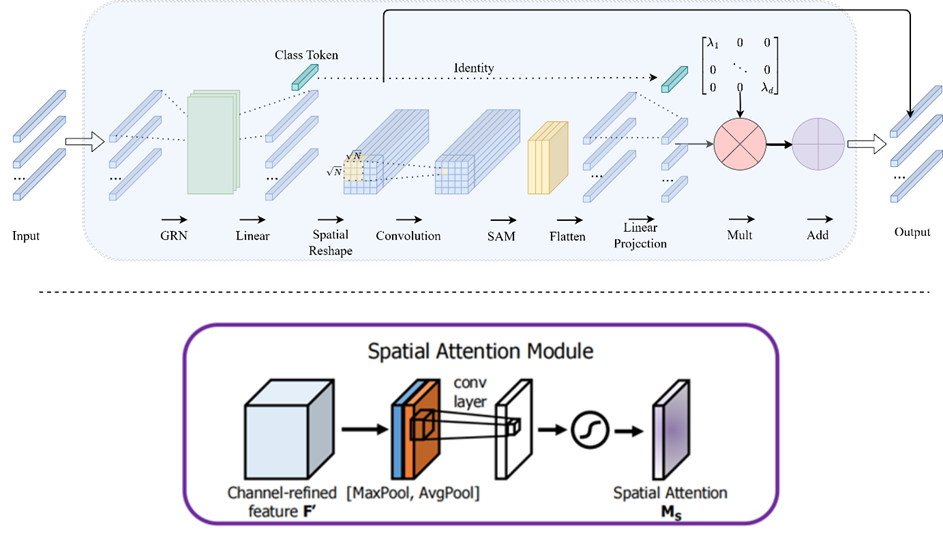}
	\caption{Diagram of a lightweight food image classification algorithm based on window multi-head attention.}
	\label{fig2-2}
\end{figure*}

As shown in Figure \ref{fig2-2}, the principle of the spatial attention mechanism involves generating a spatial attention map through spatial relationships of features. Unlike channel attention that focuses on channel-wise information, spatial attention emphasizes complementary spatial information. To compute spatial attention, the module first applies average pooling and max pooling operations along the channel axis, then concatenates these results to create an effective feature descriptor. Pooling operations along both the channel and spatial dimensions have proven effective in highlighting informative regions. On cascaded feature descriptors, this module applies convolutional layers to generate a spatial attention map 
${{M}_{s}}\left( F \right)\in {{\mathbb{R}}^{H\times W}}$., which encodes where to emphasize or suppress features. The detailed operations are as follows:
First, through two consecutive merge operations, the channel information from the feature maps is aggregated to generate two 2D maps: $F_{avg}^{s}\in {{\mathbb{R}}^{1\times H\times W}}$ and $F_{max}^{s}\in {{\mathbb{R}}^{1\times H\times W}}$ Each channel represents the average pooling feature and the maximum pooling feature. Then, these are connected and convoluted by standard convolutional layers to produce our 2D spatial attention map. In short, the spatial attention is calculated as:

\begin{equation}
   ~~~{{M}_{s}}\left( F \right)=\sigma \left( {{f}^{\left( 7\times 7 \right)}}\left( \left[ AvgPool\left( F \right);MaxPool\left( F \right) \right] \right) \right)  \\
   =\sigma \left( {{f}^{7\times 7}}\left( \left[ F_{avg}^{s};F_{max}^{s} \right] \right) \right)  \\
\end{equation}

where $\sigma$ denotes the sigmoid function, and ${{f}^{7\times 7}}$ represents the convolution operation with a filter size of $7\times 7.$.

\subsection{Window Attention}
\begin{figure*}[!ht]
	\centering	
	\includegraphics[width=12.29cm,height=5cm]{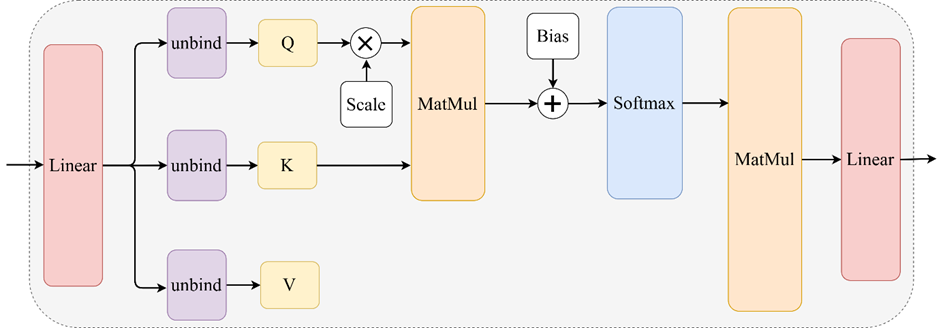}
	\caption{Window Multi-Head Attention Mechanism Diagram}
	\label{fig2-3}
\end{figure*}

As shown in Figure \ref{fig2-3}, the Multi-Head Attention mechanism is an important attention mechanism in deep learning, representing an improvement over traditional attention mechanisms. By introducing multiple parallel attention heads, the Multi-Head Attention mechanism enhances the model's ability to comprehensively capture and process input data. Through parallel processing and aggregation of results from multiple attention heads, it can capture more complex feature relationships, thereby strengthening the model's expressive power. However, since Multi-Head Attention requires simultaneous processing of multiple subspaces (i.e., multiple heads), each head necessitates independent attention calculations, leading to relatively high computational complexity. When handling large-scale data, this high computational complexity may become a performance bottleneck, limiting the model's training and inference speed. Each subspace (head) requires a separate weight matrix, significantly increasing the model's parameter count \cite{vaswani2017attention}. The rise in parameters not only escalates computational load but also amplifies storage demands. Moreover, more parameters may increase the risk of overfitting during training \cite{tay2020efficient}.

Relative Positional Encoding is an improved method for representing positional information, serving as an alternative to traditional Absolute Positional Encoding (APE). It primarily captures relative positional relationships between elements in a sequence rather than relying on their absolute positions. This method was initially introduced into self-attention mechanisms to enhance the model's generalization capability regarding sequence length and relative positions. A notable advantage of Relative Positional Encoding is its ability to reduce both the model's parameter count and computational overhead. In this approach, positional relationships (e.g., relative distances) are represented through a shared encoding function. This means that encoding for each position pair does not require individual definitions but is instead generated based on patterns of relative positional information, significantly reducing the number of parameters. Additionally, Relative Positional Encoding is independent of sequence length since it focuses on relative positions (e.g., a bias of distance i-j). Thus, parameters need only be defined for a fixed range of relative positions (e.g., [-L, L]), rather than for each specific position, further minimizing the parameters required for encoding \cite{shaw2018relative}. Due to its reliance on relative rather than absolute positions, Relative Positional Encoding exhibits stronger generalization capabilities, enabling more effective handling of variable-length inputs. This indirectly reduces unnecessary parameters introduced during training, mitigating overfitting risks.

\subsection{Window Multi-Head Attention}
The Windowed Multi-Head Attention mechanism is a solution proposed to address the high computational complexity faced by global attention mechanisms when processing large-scale images or high-dimensional feature data. Global attention mechanisms require computing attention weights between every position and all other positions, resulting in a computational complexity of O(N²), where N is the input feature length. For large-sized images or high-resolution scenarios, this complexity can lead to inefficient computations. Hence, the Windowed Attention mechanism was introduced to compute attention within local windows, thereby reducing computational costs.The core idea of Windowed Attention is to partition input features (e.g., images) into multiple local windows, with attention weights computed for features within each window. These windows can slide across the input features, gradually covering all features in the input. Within each window, the attention mechanism is applied to calculate attention weights for the elements inside, and these weights are used to perform weighted aggregation of the window's elements, yielding the output for each window. The main principles are as follows.

The mathematical formulation of the Windowed Multi-Head Attention mechanism encompasses the processes of input feature partitioning, intra-window multi-head self-attention computation, and global output integration.First, given an input feature $X\in {{\mathbb{R}}^{H\times W\times C}}$ ,where $H~$ and $~W$ denote the height and width of the feature map, respectively, and $C$ represents the number of channels, the Windowed Attention mechanism partitions the input features into multiple local windows. Each window has a size of $M\times M$ and the total number of windows is $N=\frac{HW}{{{M}^{2}}}$ .For each window, its feature representation is denoted as $X_{w}^{\left( n \right)}\in {{\mathbb{R}}^{{{M}^{2}}\times C}}$.Within each window, the dimensionality of each attention head is computed based on the input feature dimensions and the number of attention heads. Subsequently, linear transformation layers are initialized to generate Query, Key and Value:

\begin{equation}
Q=X_{w}^{\left( n \right)}{{W}_{Q}},K=X_{w}^{\left( n \right)}{{W}_{K}},V=X_{w}^{\left( n \right)}{{W}_{V}}
\end{equation}

Among them,${{W}_{Q}},{{W}_{K}},{{W}_{V}}\in {{\mathbb{R}}^{C\times d}}$ are learnable weight matrices, where $d=\frac{C}{h}$ represents the dimension of single-head attention, and $h~$ is the number of heads. Then, the dot product of the query and key is calculated to generate the attention score matrix:

\begin{equation}
\text{A}=\frac{\text{Q}{{\text{K}}^{\text{T}}}}{\sqrt{\text{d}}}+\text{B}
\end{equation}

Among them,$B\in {{\mathbb{R}}^{{{M}^{2}}\times {{M}^{2}}}}$ is the relative position bias matrix,used to capture the relative position information between different feature points within the window. Subsequently, the attention scores $A\text{ }\!\!~\!\!\text{ }$ are normalized through the Softmax function and Dropout is applied.

\begin{equation}
{A}'=\text{Dropout}\left( Softmax\left( A \right) \right)
\end{equation}

And weight the value matrix $V$ with it to obtain the output features within the window:

\begin{equation}
Z={A}'V
\end{equation}

Where $Z\in {{\mathbb{R}}^{{{M}^{2}}\times d}}$. For the multi-head self-attention mechanism, each head independently performs the above operation. Finally, the outputs of all heads are concatenated, and a linear transformation is applied followed by another Dropout to obtain the final output feature of the window.

\begin{equation}
\textit{MultiHead}\left(X_{w}^{(n)}\right) = \text{Dropout}\left(\text{Concat}(Z_1, Z_2, \ldots, Z_h) W_o\right)
\end{equation}

Among them,${{W}_{o}}\in {{\mathbb{R}}^{C\times C}}$ is the output weight matrix.

\section{Results}
\subsection{Dataset}
This paper employs two widely recognized food image datasets:Food-101 \cite{Food101} and Vireo Food-172 \cite{Food172} for model training and validation. The data preprocessing in this paper adopts a combined data augmentation method of 3-Augment, TrivialAugment, and RandomErasing.

All experiments in this paper were performed on a single NVIDIA GeForce RTX 2080Ti (11G) GPU. All algorithms were implemented using Python and based on the PyTorch framework. During the training process, all images were uniformly resized to 224×224 and fed into the network. The entire model was fine-tuned for 40 epochs, and the AdamW algorithm was used to optimize the gradients, with a batch size of 8 for each epoch. Additionally, to prevent the learning process from getting stuck in a saddle point, the cosine annealing algorithm was used to adjust the learning rate, with an initial learning rate of 7e-4 and a weight decay of 5e-2. 

\subsection{Ablation Study}
To verify the effectiveness of the modules added in the model proposed in this paper, this section conducts experiments on the AlsmViT-Large model, respectively, for the data augmentation module and the global response normalization module. The evaluation metrics are accuracy (Acc.), precision (Pre.), recall (Rec.), and F1-score (F1.). The final ablation experiment results are shown in Table 1.
\begin{table*}[htbp]
\centering
\caption{Ablation experiment results on the Food-101 and Vireo Food-172 datasets.}
\begin{tabular}{l|l|c|c|c|c|c|c}
\hline
\textbf{Dataset} & \textbf{Method} & \textbf{Params (M)} & \textbf{FLOPs (G)} & \textbf{Acc. (\%)} & \textbf{Pre. (\%)} & \textbf{Rec. (\%)} & \textbf{F1 (\%)} \\
\hline
\multirow{4}{*}{Food-101} 
& AlsmViT                 & 303.4 & 59.7 & 95.17 & 95.2 & 95.17 & 95.17 \\
& AlsmViT+WMHAM           & 227.9 & 44.9 & 95.2 & 95.22 & 95.2  & 95.19 \\
& AlsmViT+SAM             & 303.4 & 59.7 & \textbf{95.33} & \textbf{95.35} & \textbf{95.33} & \textbf{95.32} \\
& \textbf{Ours}           & \textbf{227.9} & \textbf{44.9} & 95.24 & 95.28 & 95.24 & 95.24 \\
\hline
\multirow{4}{*}{Vireo Food-172} 
& AlsmViT                 & 303.5 & 59.7 & 94.29 & 94.29 & 94.29 & 94.25 \\
& AlsmViT+WMHAM           & 227.9 & 44.9 & 94.18 & 94.19 & 94.18 & 94.14 \\
& AlsmViT+SAM             & 303.5 & 59.7 & 94.19 & 94.21 & 94.19 & 94.16 \\
& \textbf{Ours}           & \textbf{227.9} & \textbf{44.9} & \textbf{94.33} & \textbf{94.35} & \textbf{95.33} & \textbf{94.29} \\
\hline
\end{tabular}
\begin{flushleft}
\small \textbf{Note:} Bold indicates the best value for the corresponding metric.
\end{flushleft}
\end{table*}

In this context, WMHAM denotes Windowed Multi-Head Attention Mechanism, and SAM represents Spatial Attention Mechanism. This paper conducts detailed ablation experiments to analyze the performance impacts of Windowed Multi-Head Attention Mechanism (WMHAM) and Spatial Attention Mechanism (SAM) on the AlsmViT model using the Food-101 and Vireo Food-172 datasets. Experimental results reveal significant differences in model complexity, performance metrics, and applicability among different methods, highlighting trade-offs in model optimization.From the perspective of model complexity, both AlsmViT+WMHAM and the "Our" method demonstrate substantially lower parameter counts (Params) and computational costs (FLOPs) compared to the original AlsmViT model. For instance, AlsmViT+WMHAM reduces parameters from 303.4M to 227.9M and FLOPs from 59.7G to 44.9G, indicating that WMHAM enhances practical applicability in resource-constrained scenarios through computational complexity reduction. In contrast, AlsmViT+SAM maintains identical parameters and FLOPs with the original AlsmViT, suggesting that SAM does not directly alleviate computational burden, though its performance improvement on specific datasets implies potential optimization through enhanced feature extraction capabilities. This complexity reduction not only facilitates model deployment but also provides novel insights for future optimization in resource-limited environments.

In terms of performance, there are significant differences in the performance of different methods on different datasets. On the Food-101 dataset, the accuracy rate and F1 score of AlsmViT+SAM are both superior to other methods, demonstrating the potential of the spatial attention mechanism in improving model performance. By enhancing the model's ability to capture spatial information, SAM can better identify the key areas in food images, thereby improving the classification accuracy. However, on the Vireo Food-172 dataset, the performance of AlsmViT+SAM was relatively weak, with an accuracy rate of only 94.19\%, which was lower than 94.29\% of the original AlsmViT. This indicates that the effect of SAM may depend on the characteristics of the data set, and its applicability in some scenarios is limited. In contrast, the "Our" method demonstrated relatively balanced performance on both datasets. Especially on the Vireo Food-172 dataset, its accuracy rate and F1 score were both superior to other methods, showing its advantage in comprehensive performance. This balanced performance indicates that the "Our" method may combine multiple optimization strategies and be able to maintain high performance stability in different scenarios.

The ablation study reveals critical trade-offs among complexity, performance, and applicability. While WMHAM and SAM present respective advantages and limitations, the "Our" method achieves balanced optimization across multiple metrics, offering valuable references for model design. Future research could explore: Combining WMHAM and SAM to simultaneously reduce computation and enhance feature extraction; Incorporating other attention mechanisms (e.g., channel attention or hybrid attention) for performance improvement; Integrating proposed methods with advanced techniques like multimodal learning or self-supervised learning to expand application scope. These directions may provide novel approaches for advancing food image classification.

\subsection{Comparative test and analysis}
In order to further verify the performance of the proposed algorithm, this section compares the proposed method with other classical classification network models on the Food-101 and Vireo Food-172 datasets, and all experiments use the same data partition and experimental settings. The results of the comparative experiment are shown in Table 2.

\begin{table*}[htbp]
\centering
\caption{Comparison experimental results on the Food-101 and Vireo Food-172 test sets.}
\begin{tabular}{l|l|c|c|c|c|c|c}
\hline
\textbf{Dataset} & \textbf{Method} & \textbf{Params (M)} & \textbf{FLOPs (G)} & \textbf{Acc. (\%)} & \textbf{Pre. (\%)} & \textbf{Rec. (\%)} & \textbf{F1 (\%)} \\
\hline
\multirow{7}{*}{Food-101} 
& AlsmViT         & 303.4 & 59.7  & 95.17 & 95.2 & 95.17 & 95.17 \\
& MAE-L     & 303.2 & 59.7  & 92.52 & 92.53 & 92.52 & 95.51 \\
& DeiT v2-L & 303.2 & 59.7  & 94.20 & 94.22 & 94.2 & 94.21 \\
& BeiT-L  & 303.5 & 59.7  & 93.56 & 93.58 & 93.56 & 93.59 \\
& ViTamin-L& 332.4 & 72.5  & 94.86 & 94.88 & 94.86 & 94.86 \\
& ConvNeXt-XL & 348.3 & 60.96 & 92.77 & 92.80 & 92.77 & 92.76 \\
& \textbf{Ours}                 & \textbf{227.9} & \textbf{44.9} & \textbf{95.24} & \textbf{95.28} & \textbf{95.24} & \textbf{95.24} \\
\hline
\multirow{7}{*}{Vireo Food-172} 
& AlsmViT         & 303.5 & 59.7  & 94.29 & 94.29 & 94.29 & 94.25 \\
& MAE-L    & 303.3 & 59.7  & 92.75 & 92.78 & 92.75 & 92.76 \\
& DeiT v2-L & 303.3 & 59.7  & 93.70 & 93.75 & 93.70 & 93.74 \\
& BeiT-L   & 303.6 & 59.7  & 93.33 & 93.34 & 93.33 & 93.34 \\
& ViTamin-L& 332.4 & 72.5  & 94.12 & 94.14 & 94.12 & 94.13 \\
& ConvNeXt-XL & 348.3 & 60.96 & 89.45 & 89.43 & 89.45 & 89.30 \\
& \textbf{Ours}                 & \textbf{227.9} & \textbf{44.9} & \textbf{94.33} & \textbf{94.35} & \textbf{95.33} & \textbf{94.29} \\
\hline
\end{tabular}
\begin{flushleft}
\small \textbf{Note:} Bold indicates the best value for the corresponding metric.
\end{flushleft}
\end{table*}

Through comparative experiments, this article thoroughly analyzes the performance of the proposed method against current mainstream approaches in Food-101 and Vireo Food-172 datasets, with in-depth discussion of the results. The experimental results demonstrate that our method outperforms other comparative approaches across multiple evaluation metrics, including parameters (Params), computational complexity (FLOPs), accuracy (Acc.), precision (Pre.), recall (Rec.), and F1-score (F1), demonstrating significant advantages in food image classification tasks. 

Model parameters (Params) and computational complexity (FLOPs) serve as crucial indicators for evaluating model complexity and computational efficiency. The data reveal that models AlsmViT, MAE-L, DeiTv2-L, and BeiT-L maintain similar parameter sizes (around 303M) and computational costs (59.7G FLOPs), indicating comparable complexity. In contrast, ViTamin-L shows a higher parameter count (332.4M) and computational load (72.5G FLOPs), suggesting greater structural complexity. Notably, our model achieves significantly lower parameters (227.9M) and computation costs (44.9G FLOPs) while maintaining competitive performance, even surpassing other models in certain metrics.

Accuracy remains a core metric for evaluating overall model performance. On the Food-101 dataset, our model achieves 95.24\% accuracy, slightly outperforming AlsmViT's 95.17\%. Although ViTamin-L shows respectable accuracy, its higher computational cost diminishes practical value. Similar superiority is observed on the Vireo Food-172 dataset, where our model reaches 94.33\% accuracy compared to AlsmViT's 94.29\%. These results confirm our model's capability to maintain lower computational costs while achieving superior accuracy.

Comparative analysis shows that our model achieves an optimal balance between computational efficiency and performance metrics. Despite ViTamin-L's comparable performance, its higher resource demands present practical deployment challenges. ConvNext-XL shows notably inferior performance on Vireo Food-172 with similar computational costs (60.96G FLOPs) to other models, indicating poor cost-effectiveness. Our model demonstrates distinct advantages in comprehensive performance, which is particularly suitable for resource-constrained scenarios. Analysis of parameters, computational costs, and performance metrics confirms our model's excellent performance across both datasets, achieving optimal equilibrium between computational efficiency and classification capability. As visualized in Figure 6 through heatmap analysis, our model accurately identifies food categories, demonstrating robust discriminative capabilities across diverse food types.
\begin{figure*}[!ht]
	\centering	
	\includegraphics[width=12.29cm,height=5cm]{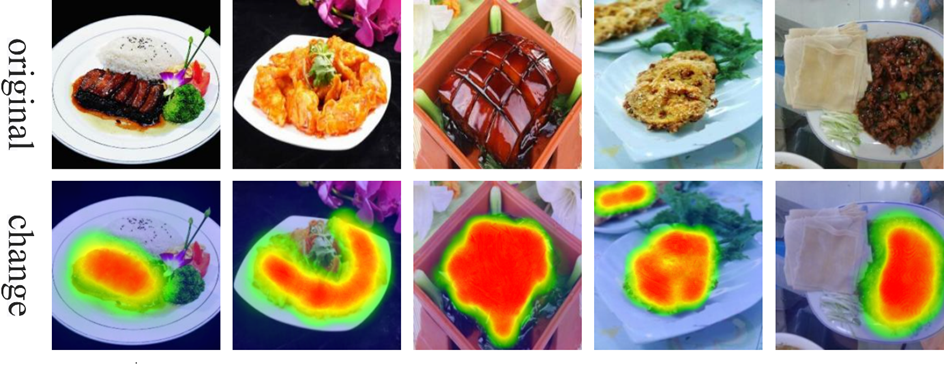}
	\caption{Visualization of classification result heatmap}
	\label{fig3-1}
\end{figure*}

\section{Conclusion}
\label{}
We propose a lightweight improvement for the Vision Transformer-based food image classification algorithm, specifically introducing the Window-based Multi-Head Attention Mechanism (WMHAM) and Spatial Attention Mechanism (SAM) to reduce model parameter size and computational complexity while enhancing classification performance. Through ablation studies and comparative experiments, we validate the effectiveness of the proposed methods. Experimental results demonstrate that the improved model achieves outstanding performance on both the Food-101 and Vireo Food-172 datasets, particularly striking a favorable balance between computational efficiency and classification performance.Specifically, the improved model exhibits significantly lower parameters (227.9M) and computational costs (44.9G FLOPs) compared to the original model (303.4M parameters and 59.7G FLOPs), while outperforming or matching the original model in metrics including accuracy, precision, recall, and F1-score. This indicates that the proposed lightweight design not only reduces computational burden but also maintains high classification performance, offering a feasible solution for resource-constrained applications.

From the results of the ablation experiment, it can be seen that the Window-based Multi-Head Attention Mechanism (WMHAM) and Spatial Attention Mechanism (SAM) exhibit varying strengths across datasets. WMHAM reduces computational complexity by partitioning input features into local windows for intra-window attention computation. However, its performance on the Vireo Food-172 dataset slightly underperforms the original model, suggesting limitations in capturing global features for complex datasets. In contrast, SAM enhances spatial feature extraction, yielding significant performance gains on Food-101, though at higher computational costs and with inconsistent cross-dataset stability. The combined approach "Our" achieves balanced performance across metrics, particularly excelling on Vireo Food-172, demonstrating effective trade-offs between complexity and classification capability.

The results of the comparative experiments further verified the superiority of the proposed method. Compared with the current mainstream food image classification models (such as MAE-L, DeiTv2-L, BeiT-L, ViTamin-L and ConvNext-XL), the "Our" model significantly reduces the number of parameters and computational cost, while outperforming or approaching other models in terms of accuracy, precision, recall and F1 score.For instance, on Food-101, "Our" achieves 95.24\% accuracy , and on Vireo Food-172, 94.33\% accuracy. The model also shows advantages in recall and F1-score, reflecting robust positive sample detection and balanced classification performance.

Overall,this study demonstrates that strategic lightweight design enables balanced performance and efficiency for food image classification, offering novel solutions for intelligent applications in resource-limited environments.

\bibliographystyle{unsrt}  
\bibliography{templateArxiv}

\end{document}